\documentclass{article}
\usepackage{ijcai24}

\usepackage{times}
\usepackage{soul}
\usepackage{url}
\usepackage[hidelinks]{hyperref}
\usepackage[utf8]{inputenc}
\usepackage[small]{caption}
\usepackage{graphicx}
\usepackage{amsmath}
\usepackage{amsthm}
\usepackage{booktabs}
\usepackage{algorithm}
\usepackage{algorithmic}
\usepackage[switch]{lineno}

\usepackage{amsfonts,amssymb} 
\usepackage{float}                  
\usepackage{subfig}                 
\usepackage{overpic}                
\usepackage{float}

\title{BET: Explaining Deep Reinforcement Learning through The Error-Prone Decisions}

\author{
Xiao Liu$^1$
\and Jie Zhao$^1$
\and Wubing Chen$^2$
\and Mao Tan$^1$
\and Yongxing Su*$^1$
\affiliations
$^1$School of Automation and Electronic Information, Xiangtan University, China\\
$^2$State Key Laboratory for Novel Software Technology, Nanjing University, China.\\
\emails
liuxiao730@outlook.com,
202121622919@smail.xtu.edu.cn,
wuzbingchen@foxmail.com,
mr.tanmao@gmail.com,
suyongxin@xtu.edu.cn
}

\begin{document}

\maketitle

\begin{abstract}
    Despite the impressive capabilities of Deep Reinforcement Learning (DRL) agents in many challenging scenarios, their black-box decision-making process significantly limits their deployment in safety-sensitive domains. Several previous self-interpretable works focus on revealing the critical states of the agent's decision. However, they cannot pinpoint the error-prone states. To address this issue, we propose a novel self-interpretable structure, named Backbone Extract Tree (BET), to better explain the agent's behavior by identify the error-prone states. At a high level, BET hypothesizes that states in which the agent consistently executes uniform decisions exhibit a reduced propensity for errors. To effectively model this phenomenon, BET expresses these states within neighborhoods, each defined by a curated set of representative states. Therefore, states positioned at a greater distance from these representative benchmarks are more prone to error. We evaluate BET in various popular RL environments and show its superiority over existing self-interpretable models in terms of explanation fidelity. Furthermore, we demonstrate a use case for providing explanations for the agents in StarCraft II, a sophisticated multi-agent cooperative game. To the best of our knowledge, we are the first to explain such a complex scenarios using a fully transparent structure.

\end{abstract}

\section{Introduction}
Deep Reinforcement Learning (DRL) has shown impressive capabilities in a variety of challenging scenarios, such as games \cite{lanctot2017unified,vinyals2019grandmaster,lin2022juewu} and robots\cite{ibarz2021train,brunke2022safe}. However, existing DRL algorithms exhibit a \textit{black-box} nature. Specifically, these models typically operate in a manner that lacks transparency, thereby obscuring the decision-making process and impeding the establishment of trust among users. The black-box nature of these models significantly constrains their deployment in domains where security and reliability are of great concern. To address this challenge, it becomes imperative to enhance the transparency of DRL agents through explanatory mechanisms, thereby facilitating broader adoption and application of DRL methods. The mainstream of interpretability research targets deep learning, while we are devoted to explaining reinforcement learning models.

Prior research has proposed methods to derive explanations for DRL. For example, saliency map-based methods \cite{paszke2019pytorch,bertoin2022look} are used to highlight the critical input features that an agent uses to make its decisions. Model approximation-based methods \cite{coppens2019distilling,liu2023effective,loyola2023towards} reveal the decision-making process behind specific actions. Reward prediction-based methods \cite{guo2021edge,cheng2023statemask} identify important time steps in an episode of interaction. In the landscape of state space, it is evident that states sharing similarities often exhibit practical correlations. For example, a majority of decisions made by the agent are straightforward and apparent \cite{liu2023effective}, with only a subset being critical and susceptible to errors \cite{coppens2019distilling}. Despite this understanding, there remains a gap in research that delves into interpreting policies by integrating these critical states. Filling this gap could significantly indicate the likelihood of the model being in error-prone situations.




We propose a novel self-interpretable model, named Backbone Extract Tree (BET)\footnote{The source code of BET can be found at [Anonymity]}, to pinpoint the decision risk of states. BET hypothesizes that states in which the agent consistently executes uniform decisions exhibit a reduced propensity for errors. To effectively model this phenomenon, BET express these states within neighborhoods, each defined by a curated set of representative states, which we call Bones. Therefore, states positioned at a greater distance from Bones are more prone to error. BET is distinguished by several key features: (i) \textit{Full transparency}. BET is meticulously designed to be well-structured, ensuring clarity and the absence of any ambiguous intermediate variables. (ii) \textit{Scalable interpretability}. The model maintains a consistent level of comprehensibility, irrespective of the complexity of the tasks it handles, thereby facilitating ease of understanding even in more intricate scenarios.


In summary, this paper makes the following contributions.

\begin{itemize}
    \item Innovative self-interpretable model. We propose BET, a novel tree-structured self-interpretable model, to accurately identify decision risks associated with various states. BET distinguishes itself through its features of \textit{fully transparent} and \textit{scalable interpretability}, facilitating an enhanced understanding of decision risks and potential error propensities in agent behavior.
    \item Outperforms existing methods. We evaluate BET on 6 different tasks, ranging from classic RL game (e.g., Lunar Lander) to sophisticated multi-agent cooperative game (e.g., StarCraft II). This indicates that the proposed BET outperforms existing self-interpretable models in explanation.
    \item Application case of interpretation. This study represents the first instance of employing a fully transparent interpretative model to interpret a complex, well-optimized policy within the StarCraft II environment. Our explanations cover error-prone states, perturbations to change the agent’s decisions, and the main behavioral patterns of the agent.

\end{itemize}

\section{Related Works}
Prior research on DRL explanation primarily focuses on associating an agent’s action with its observation. Technically, these methods can be summarized into the following categories:

\textbf{Saliency map-based methods} comput gradients of the model's output with respect to the input \cite{paszke2019pytorch,huber2021local,bertoin2022look} and thus, highlight the critical input features that an agent uses to determine its moves. Although these methods facilitate a visual understanding of the model's attention and significance toward different inputs, the interpretability provided by saliency map-based methods does not always convey the complete reasoning behind a model's decisions.

\textbf{Model approximation-based methods} create simplified versions or surrogate models that capture essential aspects of the original model's functionality \cite{coppens2019distilling,liu2023effective,loyola2023towards,costa2023evolving}. By doing so, these approaches provide insights into the model's decision-making process and behavior. While these methods contribute to interpretability by offering simplified representations, they may not fully reveal the comprehensive reasoning and intricacies behind the decisions made by the original model.

\textbf{Reward prediction-based methods} predict rewards or returns in reinforcement learning scenarios \cite{guo2021edge,huyuk2023explaining,cheng2023statemask}. By focusing on the anticipation of rewards or future outcomes, these methods aim to interpret the model's decision-making process. While they offer a valuable perspective on the model's behavior and objectives, they introduce new black-box processes, thus rendering the reasoning behind decisions less explainable.

Different from the previous works, our approach ensures transparency in generating high-level explanations, such as decision risk and representative behaviors. Moreover, our method shares both input and output characteristics with DRL, enabling the use of imitation learning to generate rules consistent with DRL's decision-making process.

\begin{figure*}[htbp]
	\centering
	\subfloat[]{\includegraphics[width=.80\columnwidth]{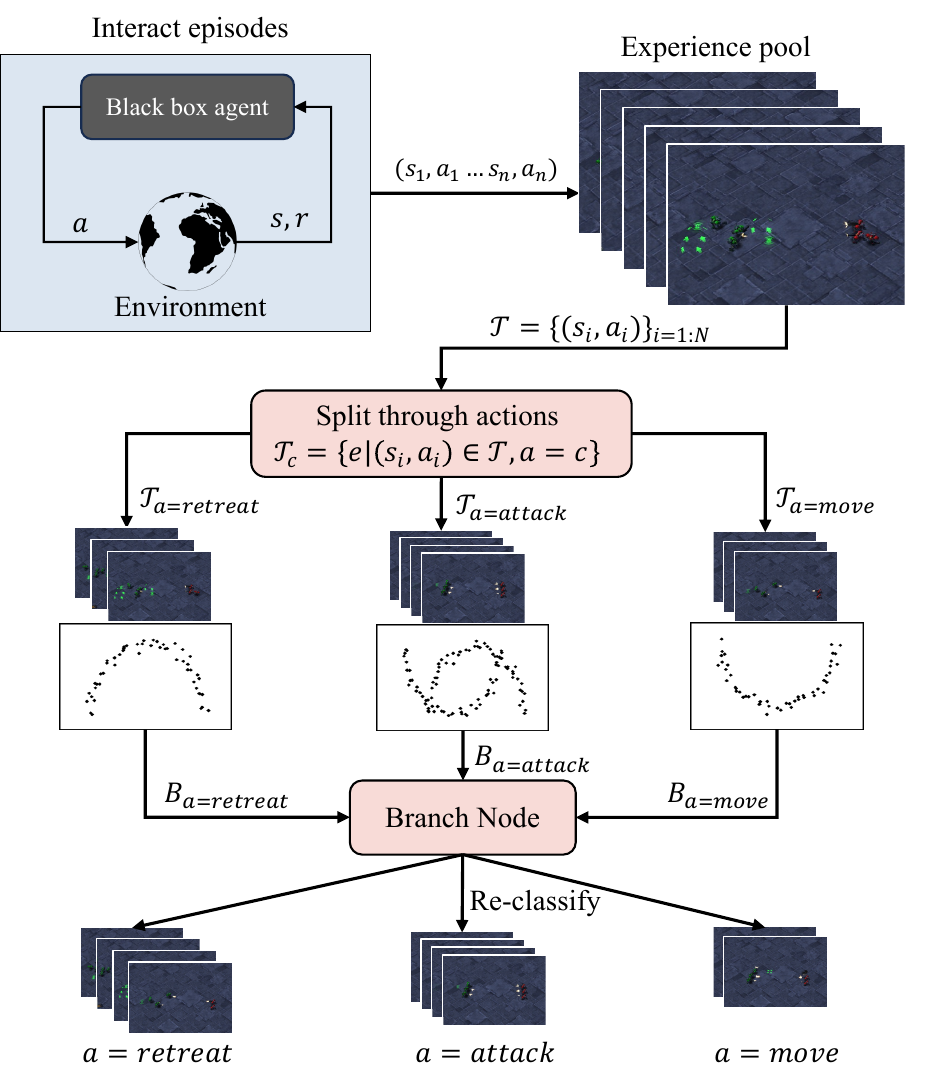}}\hspace{50pt}
	\subfloat[]{\includegraphics[width=.90\columnwidth]{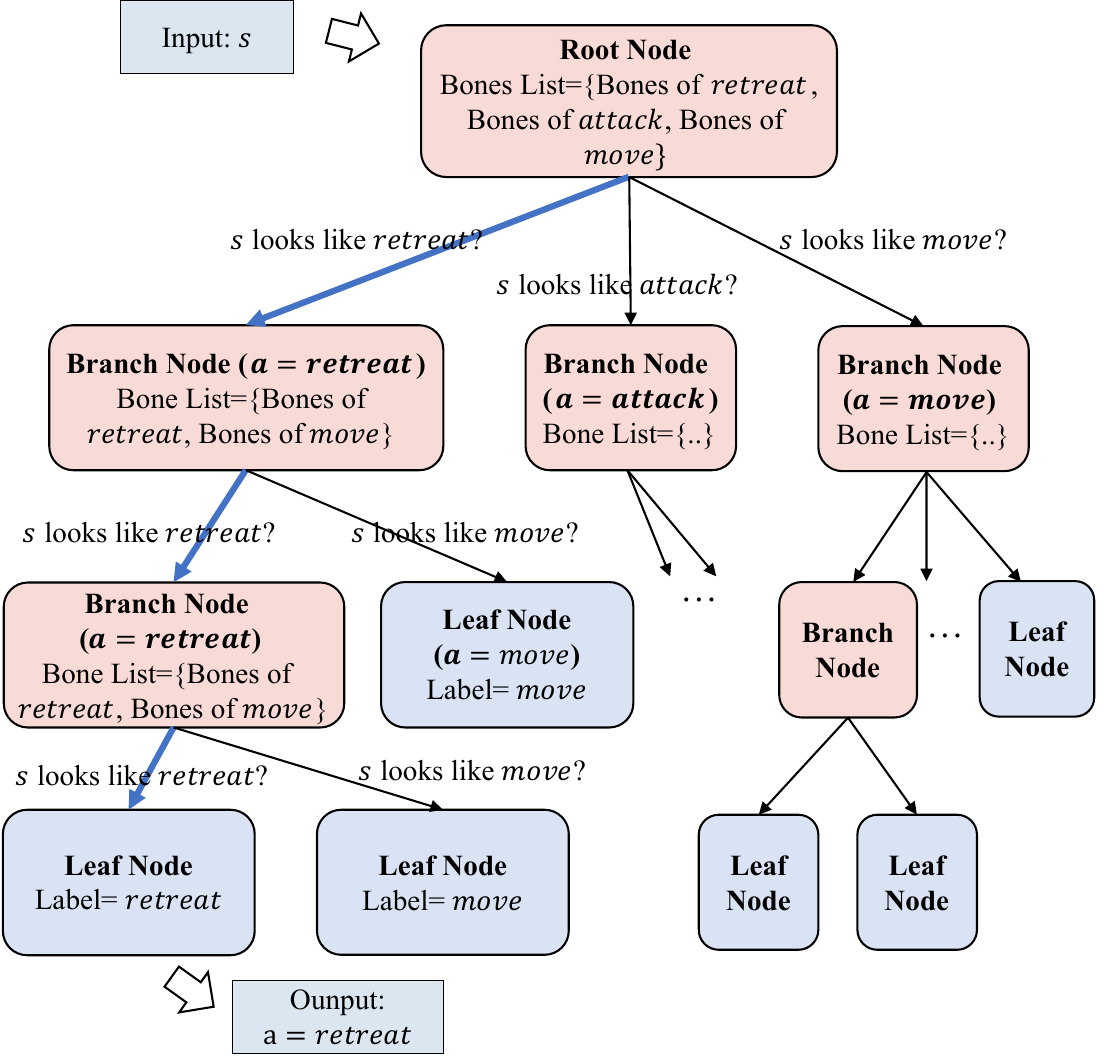}}\\
	\caption{Schematic illustration of BET: (a) Processes for building a branch of BET. (b) An intuitive case for inferring with a BET.}
    \label{fig_structure}
\end{figure*}

\section{Key Technique}

\subsection{Problem Setup}\label{ProblemSetup}
Consider a DRL game where an agent is trained using value-based \cite{nachum2017bridging} or policy-based \cite{shakya2023reinforcement} methods. Our work aims to design a novel self-interpretable model structure that reveals the decision-making process while estimating the risk of states and actions. For practical interpretability, we only allow access to the environment states and the agent’s actions. We assume that the action/state value function and the policy network are inaccessible. Additionally, the model's reasoning and backpropagation processes are unobservable.

Formally, considering a $T$-steps Markov decision process (MDP) and a teacher policy $\pi: S\rightarrow A$, the interactive trajectories follow the distribution of $\mathcal{X}$ as
\begin{equation}
\small
\begin{cases}
\mathcal{X}_0=\mathbb{I}(s=s_0)& t=0 \\
\mathcal{X}_t=\sum_{s'\in S}P(s',\pi(s'),s)\mathcal{X}_{t-1}(s')& t>0
\end{cases}.
\end{equation}
We collect its interactive trajectories $\mathcal{T}$ of $M$ episodes, i.e., 
\begin{equation}
\begin{aligned}
    &\mathcal{T}_{0}=\{(s_0,\pi(s_0)),...,(s_{T_0},\pi(s_{T_0}))\},\\
    &...\\
    &\mathcal{T}_{M}=\{(s_0,\pi(s_0)),...,(s_{T_M},\pi(s_{T_M}))\}.
\end{aligned}
\end{equation}
Then, we split the sequence into discrete state-action pairs and construct the experience pool as $\mathcal{T}=\{(s_i,\pi(s_i))\}_{i=1:\sum_{m=0}^{M} T_m, s\in \mathcal{X}}$. We use $e=(s, a)$ to represent a specific sample in the experience pool. The method we propose follows the classical Interpretable Policy Distillation (IPD) \cite{liu2023effective} framework, which involves fitting a self-interpretable model using supervised learning to minimize the validation loss of the model, i.e.,
\begin{equation}
    \mathcal{L}_{val}=\sum_{s\in S} P(s| \mathcal{X}) \mathbb{I}[\pi(s)=\hat{\pi}(s)].
\end{equation}
Our goal is to design a self-interpretable model based on the sensitivity of states and actions. The aim is to ensure that this self-interpretable model is as consistent as possible with the native DRL-based agent and thus, leveraging the inherent interpretability of the model to explain the agent's behavior.

\subsection{Straightforward Solutions and Limitations}
The most straightforward approach to identifying sensitive states/actions is to use the gradient of the model as indicators and pinpoint the input features that are sensitive to decision-making. However, since we do not assume the availability of the networks, this method is not applicable to our problem. Moreover, recall that our goal is to capture sensitivity across the entire state space, whereas gradients only indicate the sensitivity for a specific sample. Another straightforward approach is to leverage IPD techniques \cite{liu2023effective} to fit a classic decision tree model, e.g., C4.5 \cite{aldino2020decision} and CART \cite{ghiasi2020decision}, from the trajectories. However, as we show later in Appendix A1, classic decision tree models partition tree branches based on sample impurity indicators (\textit{Entropy} or \textit{Gini}). This kind of approach lacks practical meaning, and thus, the generated interpretations lack scalability. As the depth of the decision tree increases, the difficulty in understanding the model significantly escalates. Shallow trees are regarded as interpretable, whereas deep trees are not \cite{lipton2018mythos,molnar2020interpretable}.

A more realistic method is to design a prediction model based on the agent's sensitive state regions. This model refines the depiction of sensitive areas through a hierarchical structure and then imitates the agent's behavior based on the trajectories generated by the model. Each layer of this prediction model holds explicit significance, allowing individuals to intuitively understand the model's sensitive decision-making regions and gain explanations for the agent's behavior.

\subsection{Explanation Model Design of BET}\label{sec_BET_desion}
In this work, we designed a novel self-interpretable model by adopting the perspective of sensitive decisions, named BET. At a high level, BET hypothesizes that states in which the agent consistently executes uniform decisions exhibit a reduced propensity for errors. To effectively model this phenomenon, BET encompasses these states within neighborhoods, each defined by a curated set of representative states. Therefore, states positioned at a greater distance from these representative benchmarks are more prone to error. Bet represent the non-linear distributions of sensitivity regions by combining multiple such neighborhoods in a layer-wise manner. To vividly describe this concept, we refer to these representative samples as \textit{Bones}, and the layer-wise stacking of these representative samples as the \textit{Backbone}.

\paragraph{Overview.} The sensitivity of the state space implies that the model operates in a situation where it is prone to errors. Though it is difficult to depict a sensitive state space directly, certain regions of state space are inert, where the decision-making is decisive and safe. We introduce an intuition suggesting that the model is inclined to consistently opt for the same action within inert state regions. This intuition becomes apparent in trajectories where samples of a specific class are observed, revealing a concentration of these samples within the inert state space. We can determine sensitive state regions by characterizing these inert state regions. As illustrated in Fig. \ref{fig_structure}(a), our core idea is to generate a representative sample, namely the Bone. Technically, the neighborhood of the Bone represents an inert state space, where the closer a state is to the Bone, the more decisive the decision-making. To depict the sensitive regions accurately, we construct a hierarchical tree-like structure of Bones. Specifically, the BET is a multiway tree with $\mathcal{M}$ branches, each branch node consisting of $\mathcal{N}$ Bones. $\mathcal{M}$ represents the number of classes in the samples (i.e., the number of actions), with Bones of each class stored in order within the child branches. Consequently, the branches of the BET hold explicit meanings, as each node serves as a specialized identifier for a specific class. Intuitively, the BET can be envisioned as a backbone composed of multiple layers of Bones, where shallow branch nodes provide an overview of the sensitive state space, and deeper branch nodes precisely depict specific areas within that space.

\paragraph{Branch nodes.} Bones define a neighborhood within which the agent's decisions are obvious. At a high level, for a specific class (action), our method identifies regions where experience points tend to aggregate by clustering the samples. See Fig. \ref{fig_structure} (a) for an intuitive demonstration.

Given an experience pool $\mathcal{T}=\{(s_i,a_i))\}_{i=1:N}, N=\sum_{m=0}^{M} T_m$, we split $\mathcal{T}$ base on the classes as 
\begin{equation}
    \mathcal{T}_{c}=\{e|(s_i,a_i)\in \mathcal{T}, a=c\}, c=1:C,
\end{equation}
where $e=(s,a)$ and $C$ denotes the number of feasible actions. Then, to formalize the representation of an inert state space, we perform $\mathcal{N}$-center clustering separately for each subset $\mathcal{T}{c,n=1:\mathcal{N}}$. A cluster represents an area where similar samples are aggregated. We compute the Bones $B$ using
\begin{equation}
    B_j=\frac{\sum_{i=1}^N\mathbb{I}(a_i=j)e_i}{\sum_{i=1}^N\mathbb{I}(a_i=j)}, e\in \mathcal{T}_{c,n=1:\mathcal{N}},
\end{equation}
where $\mathbb{I}(\cdot)$ is the indicator function. At this point, the neighborhoods of Bones ($B$) represent the regions within the cluster where decisions are less prone to errors.
Finally, the union of the neighborhoods of Bones represents the inert state regions. When $\mathcal{N} > 2$, the model can represent a nonlinear inert state region. In summary, Bones within any given branch node can be formalized as $\bigcup (B_{c,j}), c=1:\mathcal{M},j=1:\mathcal{N}$.

\paragraph{Tree-like backbone.}
While an individual Bone-based branch node can roughly outline the region where the agent is prone to make mistakes, it does not provide precise sensitivity information. Therefore, we designed a tree-like structure, named Backbone, based on stacked branch nodes to further fine-tune sub-regions. Recall that each branch node has $\mathcal{M}$ sub-branches, each containing $\mathcal{N}$ Bones. We designed a filter based on Bones to ensure that each sub-branch learns a specific distribution for a class. In other words, our goal is to filter samples as much as possible into their corresponding sub-branches. For example, samples related to action $i$ are classified into the $i$-th sub-branch. Formally, the probability of taking the $i$-th branch is
\begin{equation}
    p_i(\mathbf{s})=\mathbb{I}[i=\arg\min_c\mathbb{E}_{i=1}^N d(\mathbf{s},\mathbf{s}_{B_{c,j}})],
\end{equation}
where $d(\cdot)$ is the distance function. This model is a hierarchical mixture of experts, but each expert is actually a ``bigot'' who does not look at the data after training. Consequently, they always produce the same distribution. The model learns a hierarchy of filters used to assign each example to a specific bigot, considering a particular path probability. Additionally, each bigot learns a simple, static distribution over the possible output classes $i$.

\paragraph{Interpretable branch.} 
Traditional decision trees typically branch to reduce the impurity of the dataset, whereas the branches do not correspond directly to the meaning of the output classifications. Therefore, interpreting the model becomes more challenging as decision trees grow larger. To address this issue, branches in BET correspond one-to-one with the classification of samples, i.e., $\mathcal{M}=C^{(b)}$, where $C^{(b)}$ represents the number of classes under branch $b$. Specifically, the number of branches in BET is dynamic. Given any branch node $b$, the number of branches depends on the number of class values observed in the training samples at that node. Since each branch represents a bigot classifier based on Bones of a specific class, the branches have clear meanings, determining the possible actions corresponding to the input state and guiding them into sub-branches related to a particular action. 

Similar to classic self-interpretable structures \cite{charbuty2021classification,xiong2021study}, the decision branches of BET are fully transparent and thus, provide a structured and hierarchical decision-making process that avoids to use ambiguous intermediate variables. This posterior inference process, aligned with human thinking, ensures BET's interpretability is as robust as that of traditional decision trees. Furthermore, different from the existing tree-based algorithms that obscure the association between branches and predicted labels, BET follows a process from broad identification to fine-grained recognition. Each branch node directly associates input features with potential labels. Therefore, as the number of layers increases in BET, its interpretability does not significantly diminish. Any node along the propagation path directly elucidates the relationship between the sample and the label.

\subsection{Parametric Optimization and Posterior Inference}

\paragraph{Parametric optimization.} 
The optimization process in BET revolves around computing the Bones for all branch nodes. In each branch, assuming the input data from the dataset is denoted as $\mathcal{T}\sim\mathcal{X}$, each Bone serves as the centroid of a cluster. BET aims to minimize the distance between all samples and their respective centroids. First, the cluster sum of distances can be formalized as
\begin{equation}
    CSS=num(\mathcal{T}_i) \mathcal{N} \cdot \mathbb{E} d(s_{e_{i,n}},s_{B_n}), a_{e,i}=a_{B_n}.
    \label{eq.CSS}
\end{equation}
where $num(\mathcal{T}_i)$ is the number of samples in subset $\mathcal{T}i\in \mathcal{T}$. BET splits the dataset into $\mathcal{M}$ subsets according to sample labels. Within a Branch node, it learns the Bones for all these subsets. When considering the entire clustering process, the total sum of squares results from the accumulation of all sums of squares within clusters, i.e., $\sum_c^{\mathcal{M}^{\ell}}CSS_c$, where $\mathcal{M}^{\ell}$ is the sub-node number of branch node $\ell$. In summary, we train the BET using a loss function that seeks to minimize the cross-entropy between each branch node, weighted by its path probability. For a single training case with input experience pool $\mathcal{T}$ and target distribution $\mathcal{X}$, the cost is
\begin{equation}
    \mathcal{J}=\sum_{\ell \in BranchNodes} P^{\ell}(\mathcal{\mathcal{T}})\sum_c^{\mathcal{M}^{\ell}}\log CSS_c,
    \label{eq_costfunction}
\end{equation}
where $P^\ell(\mathcal{T})$ is the probability of arriving at leaf node $\ell$ given the input $s$. The optimization process in BET is geared towards the iterative refinement of Bones across branch nodes. Its objective is to continually update the centroids of clusters, aiming to minimize the distance between samples and their corresponding centroids. This iterative refinement significantly enhances the model's ability to represent and understand intricate decision boundaries within the input space. By fine-tuning these centroids, BET augments its capacity to discern subtle variations and complex relationships among data points, thereby reinforcing its capability to make informed and accurate decisions based on the learned representations.

\paragraph{Posterior inference.}
BET is a prediction model that determines the posterior probability $P(a=i|\mathbf{s})$ for different classes (actions) $i$ given a specific input state $\mathbf{s}$. Fig. \ref{fig_structure} (b) demonstrates an intuitive example of inferring using a BET. At a high level, each branch within the BET functions as a miniature classifier, roughly partitioning samples into distinct sub-nodes that are subsequently identified by the sub-node. Each subtree acts as an expert classifier for a particular label. In instances where a branch node contains only one sub-node, it signifies that the model confidently ascertains a label for the given sample.

Formally, given a state $s\sim \mathcal{X}$, at the root of a BET, each class $i=1:\mathcal{M}$ contains a set of Bones $B_{c,j}$, representing feature collections specific to different labels. Utilizing a similarity metric, the $s$ associated with class $c$ can be calculated via $\mathbb{E}_{\ell=\ell_{root}} d(s, B_{c,j})$. Subsequently, these distances are transformed into similarity measures. Employing a Gaussian kernel function to represent the degree of similarity between the input state and Bones as $sim(\mathbf{s}, B_{c,j}) = \exp\left(-\frac{d(\mathbf{s}, B_{c,j})^2}{2\sigma^2}\right),$ where $\sigma$ is a parameter controlling the width of the Gaussian kernel. Utilizing these similarities, calculate the unnormalized probability $u_i(\mathbf{s})$ for each class as
\begin{equation}
     u_i(\mathbf{s}) = \frac{\exp\left(\sum_{j=1}^{\mathcal{N}} sim(\mathbf{s}, B_{i,j})\right)}{\sum_{k=1}^{\mathcal{M}} \exp\left(\sum_{j=1}^{\mathcal{N}} sim(\mathbf{s}, B_{k,j})\right)}. 
\end{equation}
Normalize the probabilities to obtain the posterior probability that the input state $\mathbf{s}$ belongs to each class as
\begin{equation}
    P(a=i|\mathbf{s}) = u_i(\mathbf{s})(\mathcal{M}\cdot\mathbb{E} u(\mathbf{s}))^{-1}.
\end{equation}
This inference process iterates recursively through the hierarchical structure of the BET, traversing from the top-level branches downwards. At each node, compute the similarity between the input state and the corresponding Bones, transform distances into similarities using the Gaussian kernel function, and ultimately compute the posterior probabilities for the input state belonging to each class. These posterior probabilities can aid in determining the most probable action selection.

\begin{table*}[]
\caption{The average rewards and win rate in different tasks. The win rate is only for StarCraft II environment.}
\begin{tabular}{|c|c|cccccccc|}
\hline
\textbf{Tasks}                                                              & \textbf{\begin{tabular}[c]{@{}c@{}}DRL\\ Baselines\end{tabular}}                                    & \textbf{ID3}                                              & \textbf{CART}                                            & \textbf{KNN} & \textbf{GBDT}                                             & \textbf{DAGGER}                                          & \textbf{VIPER}                                           & \textbf{BOCMER}                                          & \textbf{BET}                                                 \\ \hline
\textbf{Predator-Prey}                                                      & -70.23                                                   & -72.81                                                   & -72.98                                                   & -70.89  & -71.08                                                   & -70.07                                                   & -70.95                                                   & \textbf{-68.56}                                          & -70.30                                                            \\ \hline
\textbf{Flappy Bird}                                                        & 24.79                                                    & -5.00                                                    & -4.93                                                    & 14.09 & 7.98                                                    & 18.09                                                    & 14.33                                                    & \textbf{24.77}                                           & 24.46                                                             \\ \hline
\textbf{Lunar Lander}                                                       & 283.33                                                   & -353.77                                                  & -309.73                                                  & -190.79 & 223.30                                                  & 25.56                                                    & -71.47                                                   & 269.22                                                   & \textbf{282.63}                                                   \\ \hline
\textbf{\begin{tabular}[c]{@{}c@{}}StarCraft II\\ 3m\end{tabular}}          & \begin{tabular}[c]{@{}c@{}}19.21\\ (94.2\%)\end{tabular} & \begin{tabular}[c]{@{}c@{}}17.42\\ (80.3\%)\end{tabular} & \begin{tabular}[c]{@{}c@{}}17.58\\ (80.9\%)\end{tabular} & \begin{tabular}[c]{@{}c@{}}17.37\\ (79.5\%)\end{tabular} & \begin{tabular}[c]{@{}c@{}}17.51\\ (79.7\%)\end{tabular} & \begin{tabular}[c]{@{}c@{}}17.36\\ (75.7\%)\end{tabular} & \begin{tabular}[c]{@{}c@{}}17.08\\ (77.2\%)\end{tabular} & \begin{tabular}[c]{@{}c@{}}17.67\\ (81.5\%)\end{tabular} & \textbf{\begin{tabular}[c]{@{}c@{}}18.50\\ (88.2\%)\end{tabular}} \\ \hline
\textbf{\begin{tabular}[c]{@{}c@{}}StarCraft II\\ 2s\_vs\_1sc\end{tabular}} & \begin{tabular}[c]{@{}c@{}}19.88\\ (95.3\%)\end{tabular} & \begin{tabular}[c]{@{}c@{}}18.76\\ (83.2\%)\end{tabular} & \begin{tabular}[c]{@{}c@{}}18.54\\ (80.4\%)\end{tabular} & \begin{tabular}[c]{@{}c@{}}18.57\\ (81.1\%)\end{tabular} & \begin{tabular}[c]{@{}c@{}}18.60\\ (82.9\%)\end{tabular} & \begin{tabular}[c]{@{}c@{}}18.42\\ (79.3\%)\end{tabular} & \begin{tabular}[c]{@{}c@{}}18.83\\ (83.9\%)\end{tabular} & \begin{tabular}[c]{@{}c@{}}18.78\\ (83.2\%)\end{tabular} & \textbf{\begin{tabular}[c]{@{}c@{}}19.72\\ (94.0\%)\end{tabular}} \\ \hline
\end{tabular}
\label{tab_rewards_win_rate}
\end{table*}
\subsection{Model Convergence}

In this section, we prove the convergence of the BET. In each episode of the update, BET re-splits the samples ($\mathcal{T}^{\ell}$) before calculating the cost $\mathcal{J}$. The cost function of BET is given as $\mathcal{J}$ (refer to Eq. (\ref{eq_costfunction})).

To verify the monotonicity of BET, we examine the change in $\mathcal{J}$ as 
\begin{equation}
\begin{aligned}
    \Delta\mathcal{J}&=\mathcal{J}_t-\mathcal{J}_{t-1}\\
    &=\frac{\sum_{\ell} P^{\ell}(\mathcal{\mathcal{T}})\sum_c^{\mathcal{M}}\log CSS_c}{\sum_{\ell} P^{\ell}(\mathcal{\mathcal{T}})\sum_{c'}^{\mathcal{M'}}\log CSS_{c'}}\\
    &=\sum_{\ell}\frac{\mathcal{M}}{\mathcal{M'}}\cdot\mathbb{E}\frac{\log CSS_c}{\log CSS_{c'}}.
\end{aligned}
\end{equation}
Where $\mathcal{J}_{t-1}$ is the previous cost, $M'$ is the previous number of sub-node, and $CSS_{c'}$ is the sum of distance in cluster $c'$. Since $M$ is the number of labels in the experience pool, which is unchanged. For any branch node $\ell$, we have: (1) $\mathcal{J}$ remains unchanged if the Bones remain unaltered. Bones are the points within a cluster that exhibit the minimum average distance. (2) When Bones change, the previous Bones no longer represent the average distances since $\mathcal{T}^{\ell}$ is re-split, whereas the new Bones are calculated to minimize the average distance among samples within the clusters. i.e., $\frac{\log CSS_c}{\log CSS_{c'}}\in \mathbb{R}^+$. The cost within clusters calculated using the previous Bones will be smaller than the cost within clusters calculated using the new Bones. Consequently, we have $\Delta\mathcal{J}\leq 0$, indicating that $\mathcal{J}$ is monotonically decreasing.

To establish the existence of a minimum value for BET, we consider $B$ as the independent variable and $\mathcal{J}$ as the dependent variable. Since $\mathcal{J}$ is the sum of costs for all clusters within each branch, it can be decomposed into
\begin{equation}
    \begin{aligned}
        &\mathcal{J}=\sum_{\ell}\sum_{c}P^{\ell}(\mathcal{T})\log CSS_c,\\
        &\ell\in BranchNodes, c=1:\mathcal{M}^\ell.
    \end{aligned}
\end{equation}
According to Eq. \ref{eq.CSS}, the cost of a specific cluster is calculated based on the distance function $d(\cdot)$. The distance between any sample point and the Bones within a cluster is a positive value, i.e., $d \in \mathbb{R}^+$, and thus, $CSS \geq 0$. Therefore, we have $\mathcal{J} \geq 0$.

Overall, considering the monotonically decreasing nature of $\mathcal{J}$ and its lower-bound characteristics, its optimization depicts a convex function, ensuring the eventual convergence of the BET.

\section{Experiments}

\subsection{Experiment Setting}

\paragraph{Baseline methods.} Recall the proposed BET is model approximation-based, we implement 7 methods of this kind. Iterative Dichotomiser 3 (ID3) \cite{charbuty2021classification}, Classification And Regression Tree (CART) \cite{charbuty2021classification}, and K-Nearest Neighbors (KNN) \cite{xiong2021study} and Gradient Boosting Decision Tree (GBDT) \cite{sun2020gradient} represent the classic fully transparent, self-interpretable models. Dataset Aggregation (DAgger) \cite{cheng2020reduction}, Verifiability via Iterative Policy Extraction (VIPER) \cite{bastani2018verifiable}, and BOundary Characterization via the Minimum Experience Retention (BOCMER) \cite{liu2023effective} represent the interpretable policy distillation frameworks that do not change the structure of self-interpretable models.

\paragraph{Test environments.}
In this section, we evaluate BET on 6 different tasks, where a foundational dataset, i.e., \textit{Moons}, is employed to visualize the ability to fit nonlinear data. \textit{Lunar Lander} \cite{brockman2016openai} and \textit{Flappy Bird} \cite{chen2015deep} represent real-time control tasks in classic scenarios. \textit{StarCraft II} \cite{samvelyan19smac} represents super hard multi-agent confront tasks that most related interpretable works avoid involved in. As for \textit{StarCraft II}, we test our method in three tasks including \textit{3m}, \textit{2s\_vs\_1sc}.

\paragraph{Agents to be explained.} Sharing our goal is to explain a DRL-based agent, as discussed in Section \ref{ProblemSetup}. Therefore, we fully train a deep learning-based model in each task except for Moons. The experiments aim to explain such well-trained models. Specifically, in the Moons dataset task, we use the generated samples as supervisory data. In Lunar Lander and Flappy Bird, a well-trained Deep Q Network (DQN) \cite{fan2020theoretical} is used as the interpretation target. For \textit{3m} and \textit{2s\_vs\_1sc} in the StarCraft II platform, a well-trained Q-Mix \cite{rashid2020monotonic} is used as the interpretation target.
\begin{figure}[H]
	\centering
	\subfloat[]{\includegraphics[width=.9\columnwidth]{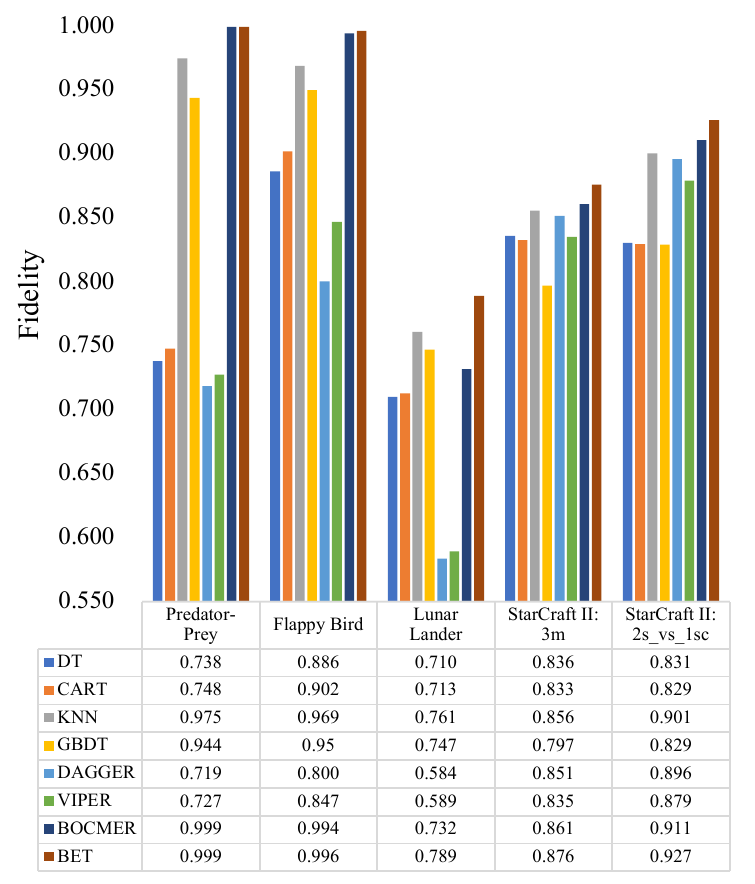}}\\
	\subfloat[]{\includegraphics[width=.45\columnwidth]{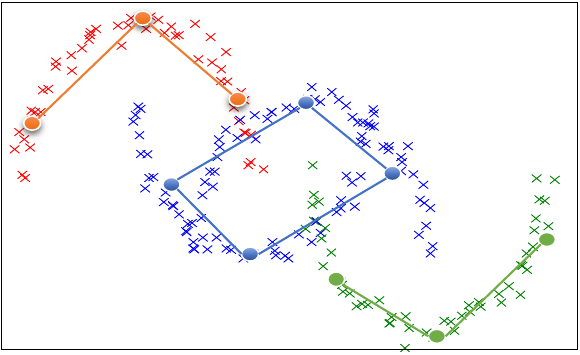}}\hspace{10pt}
	\subfloat[]{\includegraphics[width=.45\columnwidth]{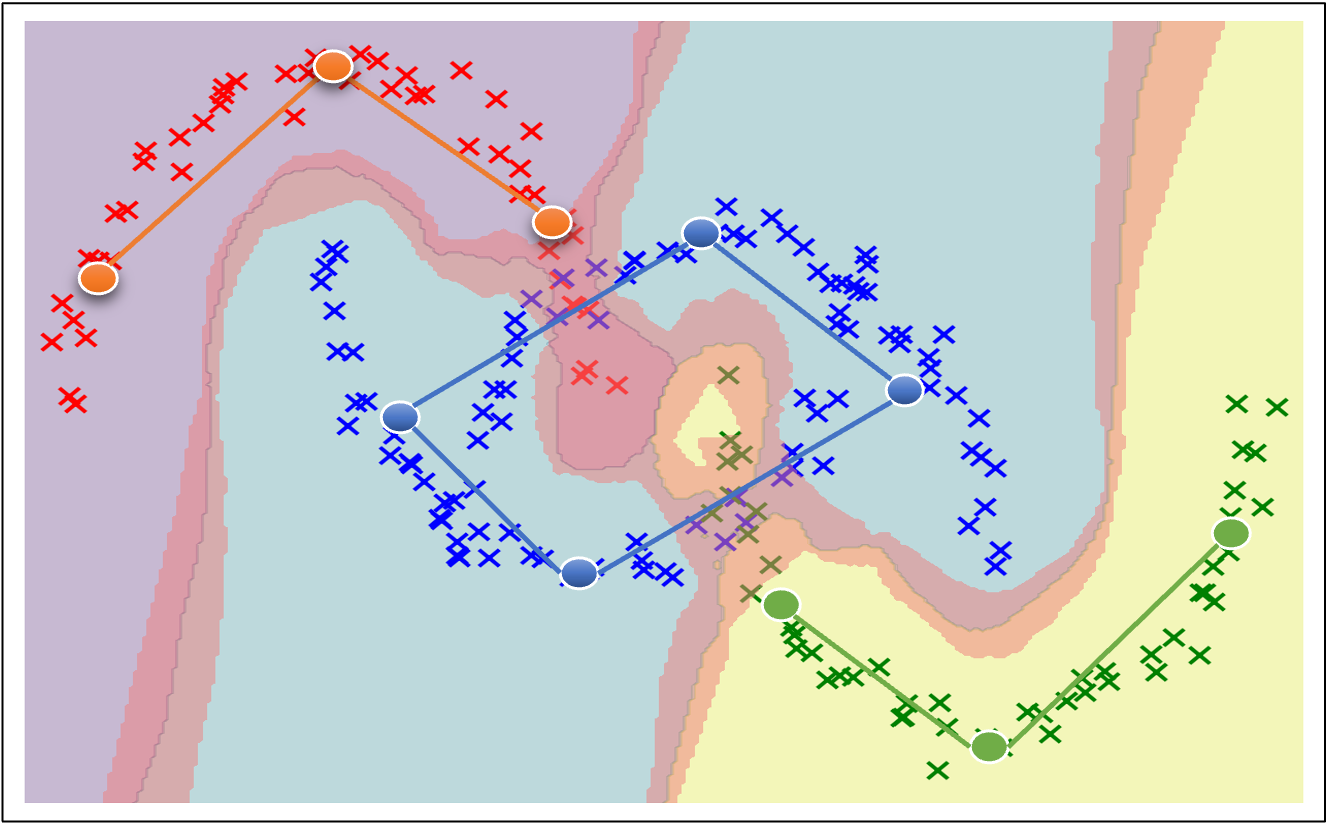}}\\
	\caption{(a) The average fidelity in different tasks. (b) Visualize bones. (c) Visualize error-prone areas (highlight in red).}
    \label{fig_visual_verification}
\end{figure}


\subsection{Experiment Results}
\paragraph{Reward and win rate.} 
Considering that the proposed method interprets by mimicking decision-making behaviors, the performance of the self-interpretable model to interact with the environment is a basic metric of the model's capabilities. Table \ref{tab_rewards_win_rate} shows the comparison results on the reward of BET against baseline alternative methods. We take the mean of test result for 200 episodes of games. First, we observe that all these methods work well in Predator-Prey and Flappy Bird, since these two games are simple in decision logic and have fewer input features. Second, for difficult tasks such as Lunar Lander and StarCraft II, the performance of comparison methods decreases significantly, indicating difficulties in generalizing to different tasks. Third, BET is not the best for simple tasks. A possible reason is that BET is designed to model the error-prone areas in the state space, which is not necessary for simple tasks. In complex tasks, where determining the boundaries of model decisions is difficult, modeling the error-prone areas yields better results.

\paragraph{Faithfulness.} 
The output of the self-interpretable model must be consistent with the target model to ensure meaningful interpretation. Figure \ref{fig_visual_verification} (a) evaluates the faithfulness of the proposed method against models based on the baseline framework. We take the mean of 100 times of test, and each record the fidelity of 1,000 decisions. The experiment results show that the proposed BET maintains the highest level of fidelity across all tasks. As discussed in section \ref{sec_BET_desion}, BET always selects the safest sub-branch at each layer of inference, reducing the possibility of errors. The results reveal that the model based on BET outperforms the baseline models in faithfulness, thereby minimizing the probability of interpretation failure.

\begin{figure*}[htbp]
	\centering
    \includegraphics[width=2\columnwidth]{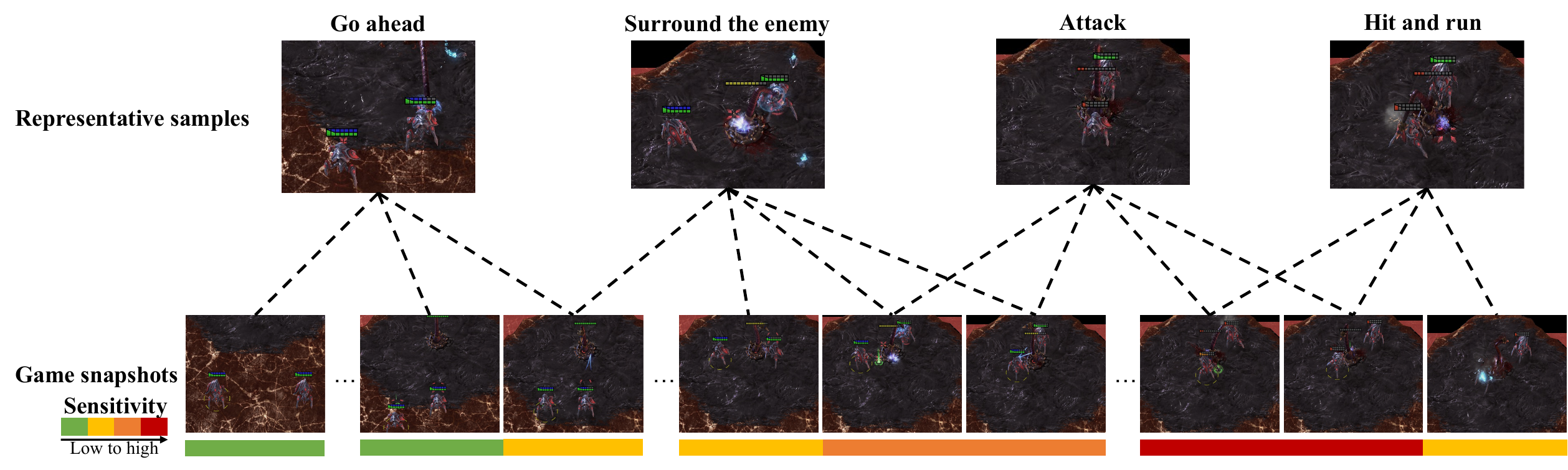}
    \caption{Agent's behavior and error-prone decisions.}
    \label{fig_behavior}
\end{figure*}

\paragraph{Visual verification.} 
To verify that BET works as expected, we visualize Bones in a two-dimensional Moons dataset. The dataset is designed with three classes, each with 100 data points. Recall that we mentioned in Section 2.1, the Bones are representative samples within a class. Figure \ref{fig_visual_verification} (b) demonstrates that Bones are generalizations of strategy in the state space. Figure \ref{fig_visual_verification} (c) visualizes the error-prone areas (highlighted in red), which represent the state regions where the model is prone to making decision errors. It can be observed that the risk areas are concentrated near the decision boundaries, which aligns with our intuition.

\section{Use Cases of Interpretation}
In this section, we demonstrate a use case of interpreting a well-trained policy in the $2s\_vs\_1sc$ task of StarCraft II. In $2s\_vs\_1sc$, ``2$s$'' represent two stalkers, ``1$sc$'' represents a spine crawler. Since the stalker is weaker than the spine crawler, to win the game, the two stalkers must alternate taking damage. In comparison to traditional tree-based models, BET offers a richer form of interpretation. 

\paragraph{Agent's Behavior}
Figure \ref{fig_behavior} showcases snapshots of the Bones of the target agent. Recall that the Bones are representative samples of a specific action. By observing the snapshots of Bones, we find that the agent wins a game using four sub-strategies: \textit{go ahead}, \textit{surround the enemy}, \textit{attack}, and \textit{hit and run}. This helps us understand several behaviors of an agent by combining the results of critical snapshots: i) the two stalkers \textit{go ahead} when they can not see the enemy. ii) When the two stalkers observe the spine crawler, they \textit{surround the enemy} and then \textit{attack}. iii) When the health of a stalker is low, it do \textit{hit and run} to escape the attack of the spine crawler. Therefore, by examining the Bones in combination with playtime, we can discover the agent's behavior patterns.

\paragraph{Error-prone decisions}
Figure \ref{fig_behavior} showcases some episode snapshots of the target agent together with the error-prone estimation extracted by our method. As observed from the results, first, decision sensitivity varies at different time steps; the \textit{go ahead} strategy is quite apparent, followed by \textit{surround the enemy}. The challenge for the agent lies in distinguishing between the \textit{attack} and \textit{hit and run} strategies, indicating that the conditions for these strategies are similar and thus, more significant throughout the episode. From a human perspective, when a stalker's health is low, it should escape the enemy's attack, allowing another stalker with higher health to take some damage. Consequently, the results indicate the error-prone steps, which align with human cognition.

\begin{figure}
    \centering
    \includegraphics[width=.5\textwidth]{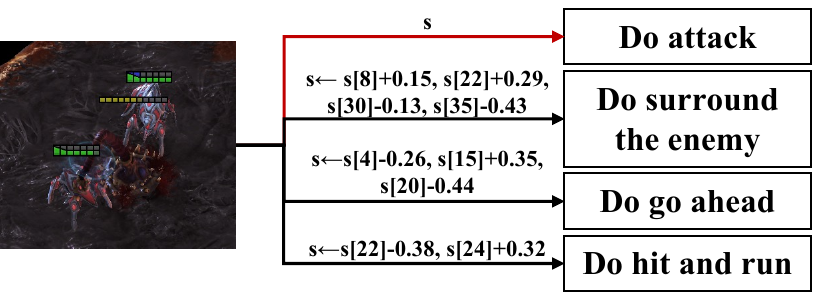}
    \caption{Perturbation on a specific state to mislead the agent.}
    \label{fig_perturbation}
\end{figure}

\paragraph{Perturbation to change agent's decision.}
Recall each branch node of the proposed BET roughly outlines the region where the agent is prone to make mistakes via Bones. BET can estimates a perturbation that leads the agent to make a wrong decision, without prior knowledge of the agent's parameters. Thus, the minimum perturbation that would make the decision go wrong can be calculated. Technically, the Bone point can be regarded as the anchor point of a type of strategy. For any state $s$, the closer $s$ is to a certain Bone point, the more certain the model is about performing the corresponding action. Therefore, a perturbation to $s$ can be calculated to mislead the model into making a different decision. Figure \ref{fig_perturbation} demonstrates an example of changing the agent's decision from \textit{attack} to other strategies. According to the interpretable result, the decision of \textit{attack} is similar to \textit{hit and run}. Combining with Figure \ref{fig_behavior}, the difficulty of distinguishing between \textit{attack} and \textit{hit and run} leads to these error-prone states.

\section{Conclusion and Discussion}
This paper proposes BET, a novel self-interpretable model structure for interpreting deep reinforcement learning. The use of BET as an explainer reveals the sensitive states where the agent may make mistakes. Unlike existing methods, BET allows humans to understand the decision risks of the agent. Experimental results demonstrate that BET models can represent agents with high fidelity. We have also showcased through case studies that BET is capable of generating a variety of rich explanatory formats.

Our work suggests several promising directions for future research. Firstly, considering the high environmental costs in safety-sensitive domains, it is impractical to extensively sample the environment. As part of future work, we plan to explore few-shot learning methods for explanations with small sample sizes. Secondly, although BET provides intuitive explanations, it falls short in extracting the high-level intentions of strategies. A potential solution for future research could be to imbue state spaces with meaning, thereby recognizing and delineating the behavioral processes of agents. Finally, this work does not delve deeply into the error-prone states that indicate the model's flaws in the state space. Therefore, one of our future objectives is to investigate these error-prone states. This could involve identifying and reinforcing learning in these areas, thereby enhancing the overall effectiveness and reliability of the model in complex environments.

\appendix

\section*{Acknowledgments}
This project was supported by the key project of Scientific Research Fund of Hunan Provincial Education Department (No.23A0142).

\bibliographystyle{named}
\bibliography{ijcai24}


\end{document}